\documentclass[10pt,twocolumn,letterpaper]{article}

\usepackage{iccv}              

\usepackage{multirow}

\definecolor{iccvblue}{rgb}{0.21,0.49,0.74}
\usepackage[pagebackref,breaklinks,colorlinks,allcolors=iccvblue]{hyperref}

\def\confName{ICCV}
\def\confYear{2025}

\title{CLICv2: Image Complexity Representation via Content Invariance Contrastive Learning}

\author{Shipeng Liu\\
XAUAT\\
{\tt\small lsp@xauat.edu.cn}
\and
Liang Zhao\\
XAUAT\\
{\tt\small zhaoliang@xauat.edu.cn}
\and
Dengfeng Chen\\
XAUAT\\
{\tt\small chdengf@xauat.edu.cn}
}

\begin{document}
\maketitle
\begin{abstract}
Unsupervised image complexity representation often suffers from bias in positive sample selection and sensitivity to image content. We propose CLICv2, a contrastive learning framework that enforces content invariance for complexity representation. Unlike CLIC, which generates positive samples via cropping—introducing positive pairs bias—our shifted patchify method applies randomized directional shifts to image patches before contrastive learning. Patches at corresponding positions serve as positive pairs, ensuring content-invariant learning. Additionally, we propose patch-wise contrastive loss, which enhances local complexity representation while mitigating content interference. In order to further suppress the interference of image content, we introduce Masked Image Modeling as an auxiliary task, but we set its modeling objective as the entropy of masked patches, which recovers the entropy of the overall image by using the information of the unmasked patches, and then obtains the global complexity perception ability. Extensive experiments on IC9600 demonstrate that CLICv2 significantly outperforms existing unsupervised methods in PCC and SRCC, achieving content-invariant complexity representation without introducing positive pairs bias.
\end{abstract}    
\section{Introduction}
\label{sec:1intro}

\begin{figure}[t!]
  \centering
   \includegraphics[width=0.95\linewidth]{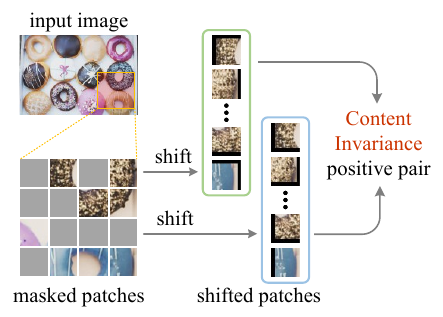}

   \caption{\textbf{Content invariant positive pairs.} \textbf{shift} denotes a randomized directional shift to patches. We redefine positive pairs via shifted patchify and combine masked entropy modeling to ensure content invariance. This effectively eliminates content interference in complexity learning.}
   \label{fig1}
\end{figure}

Image Complexity~\cite{1ic} (IC) is a central factor in computer vision tasks, directly affecting the performance of them such as image classification~\cite{1cls}, object detection~\cite{1det}, and segmentation~\cite{1seg}. High-complexity images usually contain rich textures, details, and structures, while low-complexity images are more homogeneous. This property is crucial in image quality assessment~\cite{hyperiqa}, data augmentation~\cite{1data-aug}, unsupervised learning, and adaptive computing. For example, in autonomous driving, complex scenes at night in rainy weather increase the uncertainty of perceptual models~\cite{1driving}. In medical image analysis, the complexity of lesion regions is often correlated with the difficulty of diagnosis~\cite{1medical}. Therefore, efficiently and accurately representing image complexity is a crucial challenge in computer vision.

Currently, image complexity assessment methods are mainly divided into priori-based manual feature methods and data-driven machine learning or deep learning methods~\cite{clic}. Traditional methods use manual features such as information entropy~\cite{1entropy}, edge density, and image compression ratio~\cite{1press}. Machine learning methods are dominated by supervised learning, which requires high cost to score each image's complexity and inevitably introduces subjective bias. With the development of deep learning, unsupervised contrastive learning is introduced into complexity modeling. Liu et al.~\cite{clic} proposed CLIC, an image complexity representation method based on contrastive learning, and achieved results close to supervised methods by fine-tuning on IC9600~\cite{ic9600} after pre-training on large-scale unlabeled data (Flickr-5B~\cite{flickr} and ImageNet~\cite{imagenet}).

However, we find that the pearson correlation coefficients~\cite{pcc} of the inference results of the fine-tuned CLIC vary significantly across classes, this problem that also exists in supervised learning methods (e.g., ICNet~\cite{ic9600}, ICCORN~\cite{ICCORN}, et al.). We think there are two reasons for this phenomenon. \textbf{(1) Positive pairs bias.} CLIC~\cite{clic} uses approximate IC cropping to generate views as positive samples, but the views obtained in this way may vary considerably, leading to positive pairs bias. \textbf{(2) Content sensitivity.} Even though the views can be used as positive pairs in the case of IC proximity, this will guide the model in learning the categories and object attributes in the image. Therefore, we think a well IC representation model should be insensitive to this information, meaning the model should have content invariance.

Based on these observations, we rethink CLIC~\cite{clic} positive sample selection method and propose CLICv2, a contrastive learning framework for image complexity based on content invariance constraints. \textit{Our motivation is to eliminate the interference of image content on image complexity representation and use contrastive learning to obtain a more robust IC representation.} We propose shifted patchify, a novel positive sample selection strategy that replaces direct cropping with a patch-based approach. The image is first divided into multiple patches, which undergo data augmentation followed by random directional shifts. Patches at corresponding positions are then encoded as positive pairs, ensuring content-invariant contrastive learning(\cref{fig1}). At the same time, we propose patch-wise contrastive loss which is highly coupled with shifted patchify for contrastive learning at the patch level. To further suppress image content influence, we introduce masked image modeling (MIM)~\cite{mim_survey} as an auxiliary task. However, instead of reconstructing pixels, we predict the entropy of masked patches, allowing the model to infer global complexity from the unmasked patches. This enables the model to recover missing information without relying on semantic content, preserving its ability to perceive global image complexity. The overall architecture is in \cref{fig4}. We conducted extensive experiments to validate the effectiveness of CLICv2, including comparisons with existing methods on the IC9600 dataset.In summary, we make the following contributions:

\begin{itemize}
    \item  We propose CLICv2, a contrastive learning framework that ensures content invariance by redefining positive pairs via shifted patchify and combining masked entropy modeling. This effectively eliminates content interference in complexity learning.

    \item  We propose patch-wise contrastive loss highly coupled with shifted patchify, which enables local patch-level contrastive learning to capture fine-grained complexity features while mitigating positive pairs bias.

    \item  Extensive experiments on IC9600 demonstrate that CLICv2 outperforms existing unsupervised methods in PCC and SRCC, significantly reducing positive pairs bias and improving content invariance.
\end{itemize}
\section{Related Works}
\subsection{Patch-wise Contrastive Learning}

Patch-wise contrastive learning enables the model to focus on local feature distribution. 

Wu et al.~\cite{2.1-1} analyze the dispersed distribution characteristics of cell nuclei, divide the image into multiple patches, establish two negative sample feature banks and sample and contrastive learn the features within each patch, which can capture the local features of the nuclei more efficiently. Jung et al.~\cite{2.1-2} consider the different images of an image translation task. There are differences in feature distributions between domains, and by patch-wise contrastive learning, more accurate feature correspondences can be established between images of different domains. Kınlı et al.~\cite{2.1-3} separated student model features and negative sample features by comparing the output feature maps of the student model and the teacher model at the patch level, attracting features corresponding to the features between the patches. Since the differences between normal and abnormal samples may be very subtle in industrial anomaly detection, Hyun et al.~\cite{2.1-4} can learn the feature distribution of standard samples more efficiently through patch-wise contrastive learning to improve anomaly detection accuracy. Liu et al.~\cite{2.1-5} proposed a contrastive learning method based on the structural differences of patches by calculating the source domain and target domain image semantic differences between patches and selecting positive and negative sample pairs for contrastive learning to align the feature distributions of different domains. 

Our proposed patch-wise contrastive learning method is similar to these but differs in that ours is highly coupled with shifted patchify. In addition, the negative samples are extracted from other patches of the current mini-batch for contrastive learning, so we need no memory bank for negative samples.

\subsection{Masked Image Modeling}
Inspired by Masked Language Modeling in Natural Language Processing, Masked Image Modeling (MIM) has recently been applied to learning representations for visual tasks. 

He et al.~\cite{2.2-2mae} propose MAE, which employs an asymmetric encoder-decoder architecture to learn robust visual representations by randomly masking image chunks and letting the model predict these masked chunks. Bao et al.~\cite{2.2-3beit} were influenced to propose BEiT, which splits an image into discrete visual tokens and pre-trains the image Transformer model with a masked language modeling task. SimMIM~\cite{2.2-4simmim} learns visual representations by masking predictions of overall image blocks, using a simple linear layer as a decoder that directly predicts the image blocks and avoids the complex pretraining task. DeepMIM~\cite{2.2-5deepmim} adds supervisory signals to multiple layers of the encoder so that the model can learn effective feature representations at different layers, accelerating model training and improving the effectiveness of representation learning. Wang et al.~\cite{2.2-6} proposed a local multi-scale reconstruction method, which masks the image blocks at different scales and reconstructs the local image blocks at different scales, improving the model's ability to capture the details of the image. 

\textbf{Masked Frequency Modeling.} Xie et al.~\cite{2.2-7} explored the use of frequency-domain information in self-supervised visual pretraining and proposed a masked frequency modeling method. This method improves the model's training efficiency by transforming the image into the frequency domain, randomly masking out some of the frequency components, and allowing the model to predict these masked frequency components.

\textbf{Masked Entropy Modeling (MEM).} We introduce the MIM task but redefine the reconstruction target as image entropy. Unlike pixel reconstruction, MEM enables the model to restore complexity information in masked regions without relying on semantic content, preserving its ability to capture global image complexity.

\section{Method}
\subsection{Shifted Patchify}
\label{sec3.1}

\begin{figure}[t]
  \centering
   \includegraphics[width=0.9\linewidth]{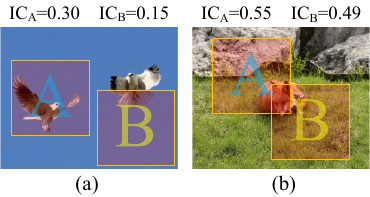}

   \caption{\textbf{View cropping of CLIC.} The size of A and B is 224 × 224. \textbf{IC} denotes the complexity score of the view.}
   \label{fig2}
\end{figure}

CLIC employs cropping to generate views as positive samples. However, this approach can lead to significant differences in content between views, increasing the complexity variation among different views of the same image (\cref{fig2}). This can negatively impact the model's learning process. While this method enhances data diversity and improves the model's ability to generalize for classification tasks, it can also create challenges in image complexity learning. Excessive variations in views may lead the model to focus on local complexity features. When views with significant complexity differences are treated as positive pairs, the model may prioritize category or object information instead of complexity. For instance, in \cref{fig2}(a), the complexity scores of the two views are 0.3 and 0.15. Using these views as positive pairs may cause the model to concentrate on similar objects rather than the underlying complexity.

We suggest that a well IC representation model should be insensitive to this information, with content invariance. For positive pairs bias, it is easy to think of using simple pixel alignment (i.e., cropping two views from the same position of the original image) in order to eliminate pixel offsets. In this way, the positive pairs lose diversity, and the pixels correspond perfectly.

Therefore, instead of performing view cropping at the image level, we consider implementing a similar process locally in the image. Inspired by the shifted window mechanism in Swin Transformer~\cite{swin}, we propose shifted patchify. However, unlike the shifted window in Swin Transformer, our approach performs random direction shifts at the patch level to break the integrity of the local region and reduce the impact of object information on the complexity representation.

\begin{figure}[t]
  \centering
   \includegraphics[width=0.8\linewidth]{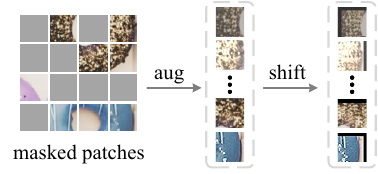}

   \caption{\textbf{Shifted Patchify.} \textbf{aug} denotes data augmentation of patches. \textbf{shift} denotes applying pixel shifts to patches in random directions. We define 8 directions: horizontal, vertical, and diagonal.}
   \label{fig3}
\end{figure}

Specifically, given an image $X$, we divide it into $G$ non-overlapping patches. Each patch undergoes a randomized directional shift (horizontal, vertical, or diagonal), disrupting local consistency and preventing the model from capturing object-level semantics. Zero-padding is applied to maintain consistent input size(\cref{fig3}). In addition, to enhance sample diversity, we introduce data augmentation (e.g., horizontal flip, color dithering, blurring et al.) before the shift operation to improve the robustness of the model. We use shifted patches as inputs to the ViT-based~\cite{vit} query encoder and key encoder. Patches at the same position are considered positive pairs with \textbf{content invariance}.

\begin{figure*}[t]
  \centering
   \includegraphics[width=0.95\linewidth]{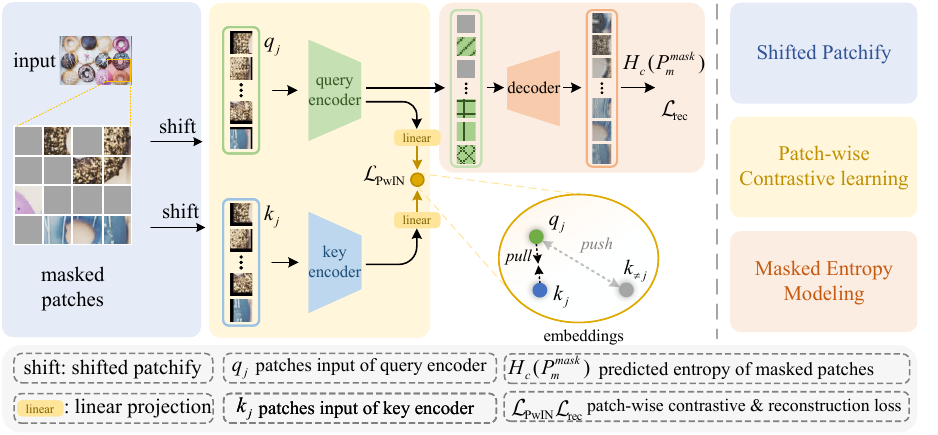}

   \caption{\textbf{Overall architecture of CLICv2.}}
   \label{fig4}
\end{figure*}

Shifted patchify solves the positive pairs bias, but the image content remains almost intact, so we consider spatially destroying the contextual semantics from the image. Similar to MAE~\cite{2.2-2mae}, we set a ratio to randomly mask a portion of the patches, destroying object and classes information in the original image. In this way, the model can \textbf{\textit{not see}} the classes and objects in the image, thus generating content invariance in the learning process. However, a serious drawback is that the model also loses the ability to model global long-distance relationships, which is also important for image complexity representation. Because the complexity of an image cannot be judged without a global perception of the image. To ensure content invariance while enhancing the ability to model global features, we introduced the masked image modeling task in \cref{sec3.3}, which enables the model to recover global complexity features based on local information.

Moreover, shifted patchify only equips the positive pairs with content invariance. To make the model learn content invariance, we further propose patch-wise contrastive learning.

\subsection{Patch-wise Contrastive Loss}
\label{sec3.2}

The goal of contrastive learning is to make the features of positive pairs closer together and negative pairs farther apart by learning the representation. MoCo~\cite{moco} compute InfoNCE~\cite{infonce} on the logits output from encoders. This is very suitable for classification tasks because it can bring similar classes closer together. However, this approach may guide the model in learning classes and object properties for image complexity representation tasks. To make the model learn content invariance, we transfer contrastive learning to the patch level and compute InfoNCE locally in the image (i.e., per patch).

\textbf{Revisiting InfoNCE $\mathcal{L}_\mathrm{IN}$.} Given a query feature $q$ and a positive sample feature $k^+$, and multiple negative sample features $k_z^-$, the loss of contrastive learning is:

\begin{equation}
    \mathcal{L}_\mathrm{IN} =-\mathbf{log} \frac{ \mathbf{exp}(q \cdot k^+ \tau)}{\mathbf{exp}(q \cdot k^+ \tau)+ \sum _z^K \mathbf{exp} (q \cdot k_z^- \tau)}
\end{equation}
Where, $\tau$ is the temperature coefficient hyperparameter.

\textbf{Patch-wise InfoNCE $\mathcal{L}_\mathrm{PwIN}$.} The input image $X$ is divided into $G$ patches ($G=P \times P$), and for each patch $P_j$, we wish to obtain its complexity-dependent feature representation through contrastive learning. To perform patch-wise contrastive learning, query and key features must be redefined. Query features are expected to be closer to positive sample features and further away from negative ones. We refer to the feature $q_j^i$ of the $j$-th patch $P_j^i$ of the $i$-th image in the mini-batch extracted by the query encoder as the query feature. The key encoder extracted by the $j$-th patch $P_j^i$ of the $i$-th image in the mini-batch with the feature $k_j^{i+}$ is the positive key feature. 

To ensure content invariance, we select negative samples from different patches within the same mini-batch, rather than maintaining a memory bank. This forces the model to distinguish complexity patterns without relying on class semantics, enhancing generalization across images. The negative key features is defined as a set $\left\{k^{i-} \right\}^N$. The contrastive loss $\mathcal{L}_j^i$ of every patch is:

\begin{equation}
    \mathcal{L}_j^i=-\mathbf{log} \frac{\mathbf{exp}(q_j^i \cdot k_j^{i+}/\tau)}{\mathbf{exp}(q_j^i \cdot k_j^{i+}/\tau)+\mathbf{exp}(q_j^i \cdot \sum \left\{k^{i-}\right\}^N/\tau)}
\end{equation}

$\mathcal{L}_j^i$ can keep patches consistent in local regions in the feature space, thus alleviating the problem of local shifts due to shifted patchify. This allows the model to consistently learn complexity information without additional feature alignment mechanisms. Similarly, this ability works equally well at the image level, but it is more difficult because the shifts are much more significant. We compute the contrastive loss separately for each patch features and average all patches to obtain the patch-wise contrastive loss $\mathcal{L}_\mathrm{PwIN}$ , which takes the following form:

\begin{equation}
    \mathcal{L}_\mathrm{PwIN} = \frac{1}{G'N} \sum_{i,j}^{G'N} \mathcal{L}_j^i 
\end{equation}
Where, $G'=G-M$ and $M$ denote masked patches, the loss is computed only for the unmasked patches. Patch-wise contrastive learning gives the model a stronger local complexity perception, but the global perception cannot be lost, which is equally important for image complexity representation~\cite{ic9600}.

\subsection{Masked Entropy Modeling}
\label{sec3.3}

The ability to model global images is important for most tasks, such as detection, segmentation, etc. Therefore, we introduce the task of masked image modeling~\cite{mim_survey} so that the model has global modeling capability. We found an essential connection between the MIM and image complexity representation tasks with theoretical analysis.

The information entropy~\cite{entropy} $H(\boldsymbol{\textbf{F}}(X)) $ can be used to reflect the complexity of the image roughly and is a measure of data uncertainty of the form:

\begin{equation}
    H(\boldsymbol{\textbf{F}}(X))=-\sum _{f \in F}p(\boldsymbol{\textbf{f}}) 
\textbf{log} p(\boldsymbol{\textbf{f}})
\end{equation}
Where, $p(\boldsymbol{\textbf{f}})$ is the probability distribution of feature $\boldsymbol{\textbf{f}}$. 

Given an image $X$, the central goal of complexity representation is to find a feature representation $\boldsymbol{\textbf{F}}$ such that the representation effectively captures the image complexity:

\begin{equation}
    C(X) \sim f(\boldsymbol{\textbf{F}}(X))
\end{equation}
Where, $f(\cdot)$ is a mapping function from the image feature representation $\boldsymbol{\textbf{F}}$ to the image complexity $C(X)$. We divide the image $X$ into $G$ patches:

\begin{equation}
    X=\left\{P_j\right\}=\left\{P_1,P_2,\ldots,P_G\right\}
\end{equation}
Where, each patch $P_j$ contributes to the overall image complexity $C(X)$. We randomly mask $M(M<G)$ patches to get $X_{mask}$:

\begin{equation}
    X_{mask}=\left\{P_j,P_m^{mask}\right\}
\end{equation}
Where, $P_m^{mask}$ denotes the masked patches, $m=0,1,\ldots,M$. $j=0,1,\ldots,G-M$. When a part of the image is masked, the information entropy $H(\boldsymbol{\textbf{F}}(X_{mask}))$ of the image decreases because the masked part cannot provide effective feature information:

\begin{equation}
    H(\boldsymbol{\textbf{F}}(X_{mask}))<H(\boldsymbol{\textbf{F}}(X))
\end{equation}

The image complexity $C(X_{mask})$ is also reduced (considering the whole image) because part of the information is obscured:

\begin{equation}
    C(X_{mask})<C(X)
\end{equation}

Thus, to represent the complexity of the whole image, the model needs to infer the context and complexity of the whole image based on the unmasked regions. The MIM task recovers the masked information by learning the contextual relations of the unmasked patches to infer the masked patches. In an information-theoretic sense, this is equivalent to increasing the conditional entropy on the masked patches through reconstruction:

\begin{equation}
    H(\boldsymbol{\textbf{F}}(X_{mask}))+H_c(\boldsymbol{\textbf{F}}(P_m^{mask})|\boldsymbol{\textbf{F}}(P_j)) \xrightarrow{} H(\boldsymbol{\textbf{F}}(X))
\end{equation}

The conditional entropy $H_c$ denotes the complexity inferred from the masked patches, given the unmasked information. The image-level reconstruction task can lead the model to learn the class, object attributes, which we do not expect. Therefore, we set the goal of the reconstruction task as the image information entropy, which makes it possible to recover the entropy of the masked patches, and thus the whole image, through the information of the unmasked patches.

We adapt the modeling objective of MIM to image entropy, which is called masked entropy modeling (MEM). In this way, MEM will not help the model learn class information. It will only enhance the complexity modeling capability. Reconstruction loss can be achieved by minimizing the following objective:

\begin{equation}
    \mathcal{L}_\mathrm{rec} =\sum_m^M \Vert H_c (\boldsymbol{\textbf{F}}(P_m^{mask}) \mid \boldsymbol{\textbf{F}}(P_j))-H(\boldsymbol{\textbf{F}}(P_m^{mask})) \Vert ^2
\end{equation}
Where, $M$ is the indexed set of masked patches. The MEM can reconstruct the information entropy of the masked portion to retrieve the global information that may have been lost in feature alignment by contrastive learning. The reconstruction network is a small ViT as the decoder. We used MEM as an auxiliary task for image complexity representation contrastive learning to obtain CLICv2 loss:

\begin{equation}
    \mathcal{L}=\mathcal{L}_\mathrm{PwIN} +\lambda \cdot \mathcal{L}_\mathrm{rec}
\end{equation}
Where, $\lambda$ is the hyperparameter, set to 2 in our experiments, \cref{fig4} shows the overall architecture of CLICv2.

\section{Experiments}
\subsection{Setup}

\begin{table}[t]
\centering
\begin{tabular}{c|cc|cc}
method                  & backbone  & sp.        & PCC↑           & SRCC↑          \\ \hline
Kyle et al.~\cite{ComplexityNet}           &           & \checkmark & 0.873          & 0.87           \\
HyperIQA~\cite{hyperiqa}                &           & \checkmark & 0.935          & 0.935          \\
P2P-FM~\cite{p2p-fm}                  &           & \checkmark & 0.94           & 0.936          \\
ICNet~\cite{ic9600}                   & ResNet18  & \checkmark & 0.947          & 0.944          \\
ICCORN~\cite{ICCORN}                  & ResNet152 & \checkmark & \textbf{0.954} & \textbf{0.951} \\ \hline
MoCo                    & ResNet50  &            & 0.759          & 0.748          \\ \hline
\multirow{3}{*}{CLIC~\cite{clic}}   & ResNet50  &            & 0.839          & 0.826          \\
                        & Swin-B    & \textbf{}  & 0.858          & 0.851          \\
                        & ViT-L/16  & \textbf{}  & 0.866          & 0.858          \\ \hline
\multirow{4}{*}{CLICv2} & ViT-S/32  &            & 0.844          & 0.838          \\
                        & ViT-B/32  & \textbf{}  & 0.865          & 0.857          \\
                        & ViT-L/16  & \textbf{}  & \textbf{0.879} & \textbf{0.87}  \\
                        & ViT-L/16* & \textbf{}  & \textbf{0.933} & \textbf{0.927}
\end{tabular}
\caption{\textbf{Comparison of supervised and unsupervised methods.} The unsupervised methods were all pre-trained on the IC1M dataset and obtained by linear probing on the IC9600. \textbf{sp.} denotes the supervised method.  \textbf{*} denotes that CLIC's dataset augmentation method was used.}
\label{tab1}
\end{table}

\textbf{IC1M dataset.} We constructed a new dataset named IC1M containing 1.5 million images. All images were randomly sampled from the Flickr-5B~\cite{flickr} dataset. If not otherwise noted, the implementation details of the experiments follow the these settings.

\textbf{Pretrain.} The CLICv2 encoders and decoder are based on ViT, where the decoder for masked entropy modeling is the smaller scale one. The update momentum of the key encoder is 0.999. Temperature coefficient of contrastive loss is 0.2. Batch size is 2048 on 4 NVIDIA Geforce RTX 3090 GPUs. Input size is 224. Mask ratio is 0.6 (\cref{fig6}). The initial learning rate is 0.003, and cosine decay adjusts the learning rate. 300 epoch are trained, and the warmup is in the first 40 epoch. The optimizer adopts AdamW~\cite{adamw}, with exponential decay of 0.9 and 0.95 and weight decay of 0.05, respectively.

\textbf{Linear probing.} We use the query encoder with frozen parameters in ICNet for linear probing. Note that ICNet uses a two-branch structure, and we only use the query encoder as a single branch. Batch size is 128 on 4 GPUs. Stochastic gradient descent (SGD) is used to optimize the model, momentum is  0.9, weight decay is  0.0001. Learning rate is 0.001. Linear probing epoch is 30. The method evaluation metrics are PCC~\cite{pcc} and SRCC~\cite{srcc}.

\subsection{Comparison with Previous Results}
We compare CLICv2 with other methods, which include supervised and unsupervised methods for evaluating image complexity (\cref{tab1}). Some unsupervised methods are not shown due to low results, the details of which can be found in ~\cite{clic}.

The experimental results (\cref{tab1}) for linear probing on the IC9600 dataset show that the supervised approach still performs well, with ICCORN obtaining a PCC and SRCC of 0.954 and 0.951 using ResNet152 as a backbone. Among the unsupervised approaches, CLICv2 achieves the highest results, with a PCC and SRCC of 0.933 and 0.927, but is still far from the best one in the supervised group. Our method CLIC v2 outperforms the original CLIC method in different configurations and achieves a performance close to that of the supervised method in the ViT-L/16* configuration. Compared to the original CLIC (ViT-L/16, PCC=0.866, SRCC=0.858), CLIC v2 exhibits higher PCC and SRCC values for ViT-L/16 configurations of the same size, validating the effectiveness of our proposed content invariance positive sample selection strategy and patch-wise contrastive learning strategy. It is also worth noting that performance of CLICv2 significantly improves as the model size increases, suggesting that it can utilize a larger model capacity to learn more robust complexity representations better.

\subsection{Ablation Study}

\begin{table}[t]
\centering
\begin{tabular}{c|ccc|cc}
\multirow{2}{*}{case} & \multicolumn{3}{c|}{IC1M pre-train} & \multicolumn{2}{c}{IC9600 linear probe} \\ \cline{2-6} 
    & SP & $\mathcal{L}_\mathrm{PwIN}$ & MEM & PCC↑  & SRCC↑ \\ \hline
(a) &    &      &     & 0.766 & 0.753 \\
(b) & \checkmark  &      &     & 0.795 & 0.786 \\
(c) &    & \checkmark    &     & 0.713 & 0.702 \\
(d) &    &      & \checkmark   & 0.789 & 0.781 \\
(e) & \checkmark  & \checkmark    &     & 0.827 & 0.817 \\
(f) & \checkmark  & \checkmark    & \checkmark   & 0.865 & 0.857
\end{tabular}
\caption{\textbf{Ablation of CLICv2. } Baseline is CLICv2 with ViT-B/32. \textbf{SP} denotes shifted patchfiy. \textbf{MEM} denotes masked entropy modeling.}
\label{tab2}
\end{table}

We set up ablation studies to verify the validity of each component (\cref{tab2}). Each set of experiments was pre-trained on IC1M and linear probed at IC9600. We observed that both SP and MEM contributed to PCC when acting alone, which validated their effectiveness. Instead, PCC and SRCC decrease when $\mathcal{L}_\mathrm{PwIN}$ alone is present. We believe this is because $\mathcal{L}_\mathrm{PwIN}$ is highly coupled with SP, which can also be verified from group (e) results. It can be found that a vast improvement is obtained compared to group (a), where the PCC and SRCC increase by 0.061 and 0.064, respectively. Enabling all modules (f) jointly achieves the highest PCC (0.865) and SRCC (0.857), surpassing all ablations.

\subsection{Content Invariance Analysis}
\begin{figure}[t]
  \centering
   \includegraphics[width=1\linewidth]{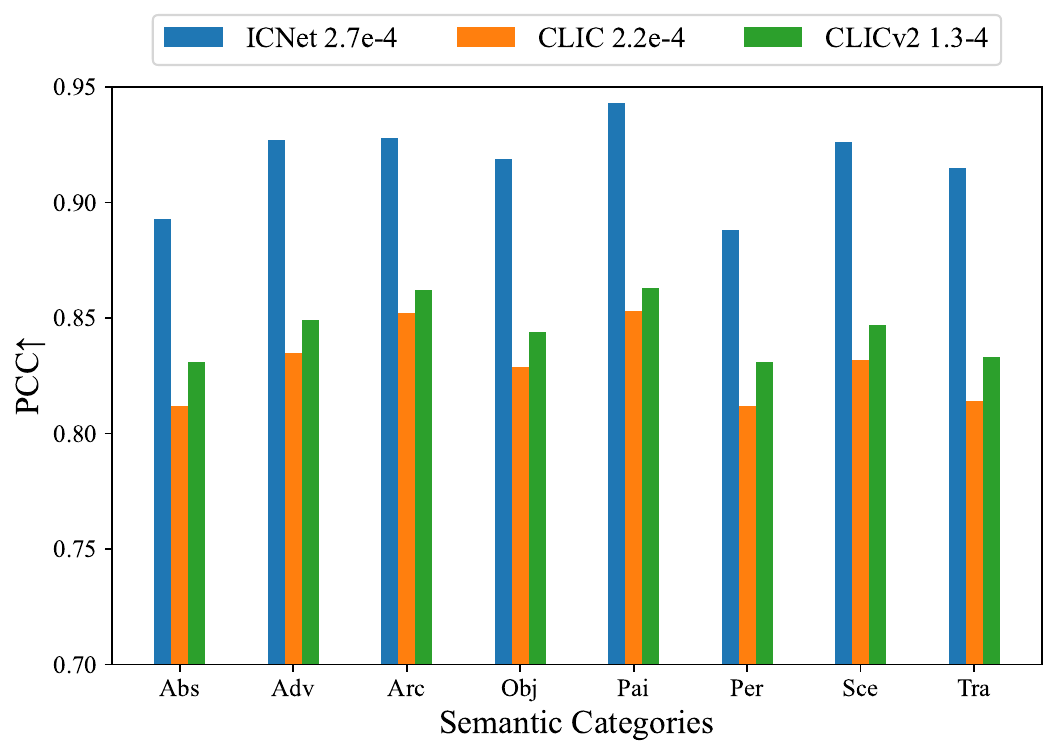}

   \caption{\textbf{PCC of semantic categories in IC9600.} Both CLIC and CLICv2 used ViT-L/16 as encoders.}
   \label{fig5}
\end{figure}

\textbf{PCC variances of semantic categories.} To verify content content invariance, we compare the inference results of ICNet, CLIC, and CLICv2 on the IC9600 test set (\cref{fig5}). It is plotted as a more intuitive bar chart according to semantic categories. Firstly, CLICv2 has higher PCC than CLIC in all semantic class results, but there is still an insurmountable gap with the supervised method ICNet. In addition, we calculated the \textit{variance} of PCC for all classes and labeled them in the legend. ICNet has the highest \textit{variance}, while CLICv2 has a substantially lower PCC \textit{variance}. The \textit{variance} reflects the homogeneity of PCC of all categories, which indicates that our method significantly eliminates class information and the model possesses content invariance.

\textbf{ImageNet top-1 acc.} We subjected the pre-trained CLICv2 to linear probing on IC9600 and ImageNet~\cite{imagenet} (IN1K). Results are in \cref{tab3}. Firstly, group (e) achieved the best results, while the top-1 acc for this group is 34.6, which is lower than all the other groups, whereas the accuracy on the IN1K classification task drops significantly ( 39.5\% to 34.6\%). Then, we find that group (d) also improves compared to group (a) due to the credit of patch-wise contrastive loss. In addition, group (c) uses image-level data augmentation, and there is a 3.3\% increase in top-1 acc compared to (b), which also shows that image-level data augmentation is friendly to the classification task, but there is almost no change in PCC and SRCC in group (c) compared to (a). Thus, the experimental results suggest that shifted patchify indeed reduces the effect of class information and thus facilitates the model to learn class-independent complexity features.

\begin{table}[t]
\centering
\begin{tabular}{c|cc|cc|c}
\multirow{2}{*}{case} & \multicolumn{2}{c|}{patchify} & \multicolumn{2}{c|}{IC9600} & IN1K      \\ \cline{2-6} 
                      & shift          & aug          & PCC          & SRCC         & top-1 acc \\ \hline
(a)                   &                &              & 0.825        & 0.816        & 39.5      \\
(b)                   &                & \checkmark   & 0.834        & 0.826        & 40.6      \\
(c)                   &                & \checkmark*  & 0.825        & 0.815        & 43.9      \\
(d)                   & \checkmark     &              & 0.830        & 0.823        & 40.0      \\
(e)                   & \checkmark     & \checkmark   & 0.854        & 0.846        & 34.6     
\end{tabular}
\caption{\textbf{Shifted patchify analysis.} \textbf{shift} is the patches shift. \textbf{aug} is the data augmentation. * means augmentation at image-level. Baseline is CLICv2 with ViT-B/32 and uses patch-wise contrastive loss. These cases are pretrained 200 epoch.}
\label{tab3}
\end{table}

\subsection{Further Analysis}
\textbf{Patch-wise vs. Image-wise.} We also compared patch-wise contrastive loss and image-wise contrastive loss (\cref{tab4}). The experimental results show that the Patch-wise contrastive learning strategy significantly improves the PCC and SRCC performance. Based on SP+MEM, the PCC is only 0.812 with image-wise contrastive learning, while it rises to 0.865 with patch-wise contrastive learning. This indicates that compared to global feature comparison, patch-wise contrastive learning can capture local complexity features more accurately and prevent the model from learning class-related information. The substantial increase in metric is also due to the high coupling with shifted patchify.

\begin{table}[t]
\centering
\begin{tabular}{cc|cc}
baseline                          & case    & PCC↑  & SRCC↑ \\ \hline
\multirow{2}{*}{\begin{tabular}[c]{@{}c@{}}CLIC v2\\ ViT-B/32\end{tabular}} & SP+MEM+$\mathcal{L}_\mathrm{IN}$ & 0.812 & 0.803 \\
                                  & SP+MEM+$\mathcal{L}_\mathrm{PwIN}$ & 0.865 & 0.857
\end{tabular}
\caption{\textbf{Comparison of Image-wise and patch-wise contrastive loss.}}
\label{tab4}
\end{table}

\begin{table}[t]
\centering
\begin{tabular}{cc|cc|c}
\multirow{2}{*}{target} & \multirow{2}{*}{$\lambda$} & \multicolumn{2}{c|}{IC9600}                                                  & IN1K          \\
                        &     & PCC              & SRCC             & top-1 acc        \\ \hline
None                    & -   & 0.827            & 0.817            & 30.9             \\ \hline
Frequency               & 1.0 & 0.831            & 0.822            & 32.6             \\ \hline
\multirow{4}{*}{Pixels} & 0.1 & 0.830            & 0.821            & 34.8             \\
                        & 0.5 & 0.835            & 0.827            & 38.1             \\
                        & 1.0 & 0.839            & 0.828            & 42.2             \\
                        &     & $\mathbf{(+.012)}$ & $\mathbf{(+.011)}$ & $\mathbf{(+11.3)}$ \\ \hline
\multirow{5}{*}{Entropy}                 & 0.1 & 0.829            & 0.818            & 31.7             \\
                        & 0.5 & 0.836            & 0.826            & 32.2             \\
                        & 1.0 & 0.846            & 0.837            & 32.9             \\
\multicolumn{1}{l}{}    & \multicolumn{1}{l|}{}   & \multicolumn{1}{l}{$\mathbf{(+.019)}$} & \multicolumn{1}{l|}{$\mathbf{(+.020)}$} & $\mathbf{(+2.0)}$ \\
                        & 2.0 & 0.865            & 0.857            & 34.6            
\end{tabular}
\caption{\textbf{Contributions of different reconstruction targets.} \textbf{target} denotes the reconstruction target, which contains frequency, pixels, and entropy, and $\lambda$ is the weight of reconstruction loss. \textbf{None} denotes no reconstruction task.}
\label{tab5}
\end{table}

\textbf{Contributions of MIM.} \cref{sec3.3} introduces the task of masked image modeling, but we modify the modeling objective to image entropy, expecting to learn information about the image's global complexity but not focus too much on the image content. We compare frequency, pixel, and entropy as modeling objectives, respectively, and the experimental results are in \cref{tab5}. The contrastive learning on different reconstruction targets reveals that the reconstruction tasks all increase the accuracy of the classification task, with the reconstruction of pixels bringing the most significant growth rate, indicating that the method tends to learn class information. When using Entropy as the reconstruction target, the growth values of PCC and SRCC ($\lambda =1$) are more significant than pixel reconstruction. In contrast, the growth value of ImageNet classification accuracy is much lower than the latter, further indicating that MEM can effectively remove the interference of class information while enhancing the global complexity modeling capability. In addition, CLICv2 achieves the best results at $\lambda=2$.

\begin{figure}[t]
  \centering
   \includegraphics[width=0.9\linewidth]{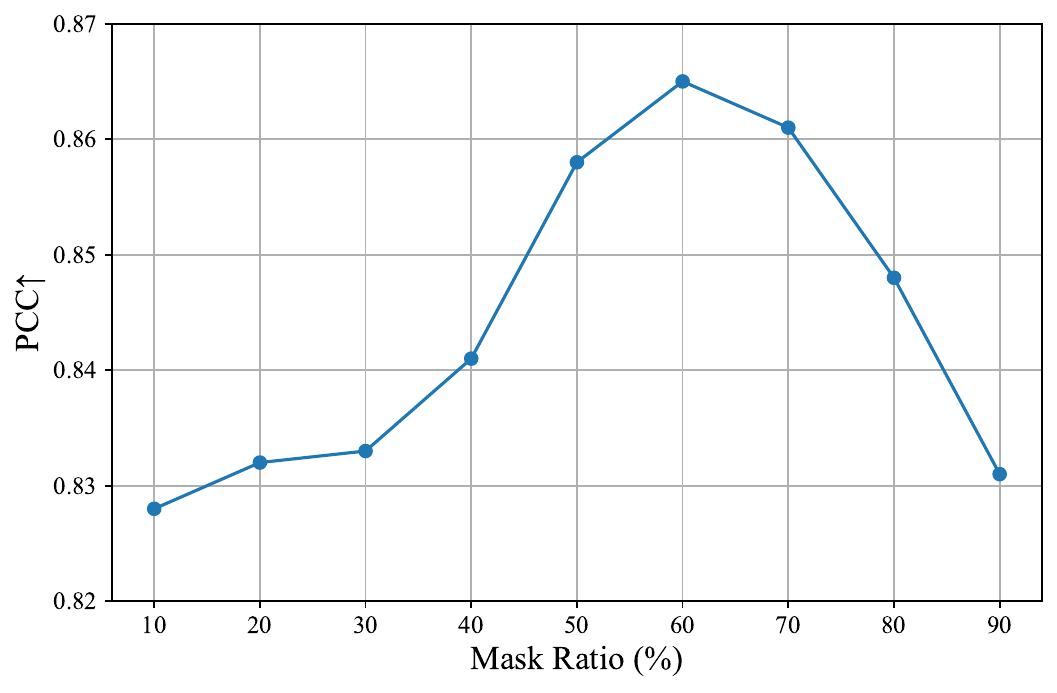}

   \caption{\textbf{Mask ratio.} Baseline is CLICv2 with ViT-B/32. $\lambda$ of reconstruction loss is 2.}
   \label{fig6}
\end{figure}

\textbf{Mask ratio. }We set different mask ratios for testing based on the CLICv2 model. Results can be seen in \cref{fig6}, the best results were achieved with a mask ratio of 0.6. This setting is also uniform for other groups of experiments.

\begin{figure}[t]
	\centering
	\begin{minipage}{0.49\linewidth}
		\centering
		\includegraphics[width=1\linewidth]{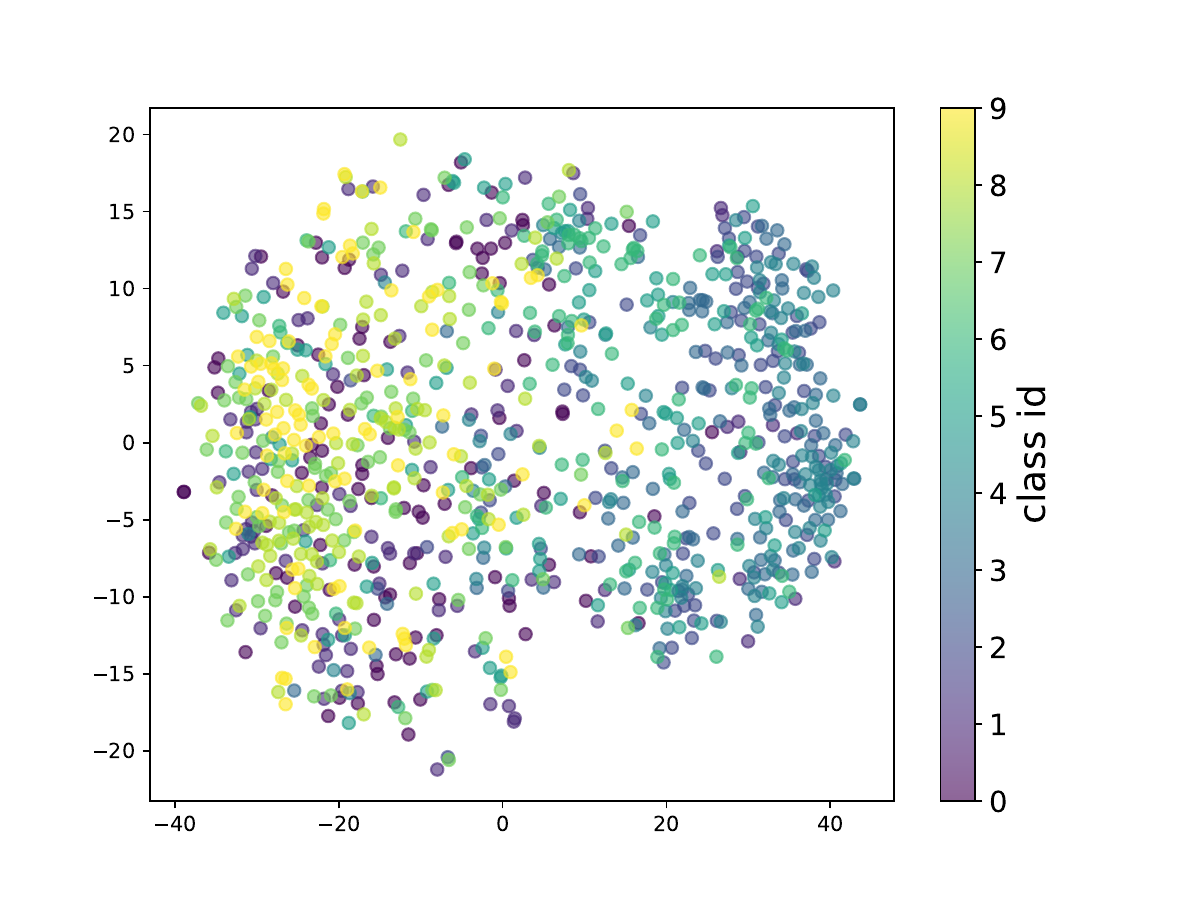}
        \subcaption{Class Coloring \\ CLIC}
		\label{figure1_1}
	\end{minipage}
	\begin{minipage}{0.49\linewidth}
		\centering
		\includegraphics[width=1\linewidth]{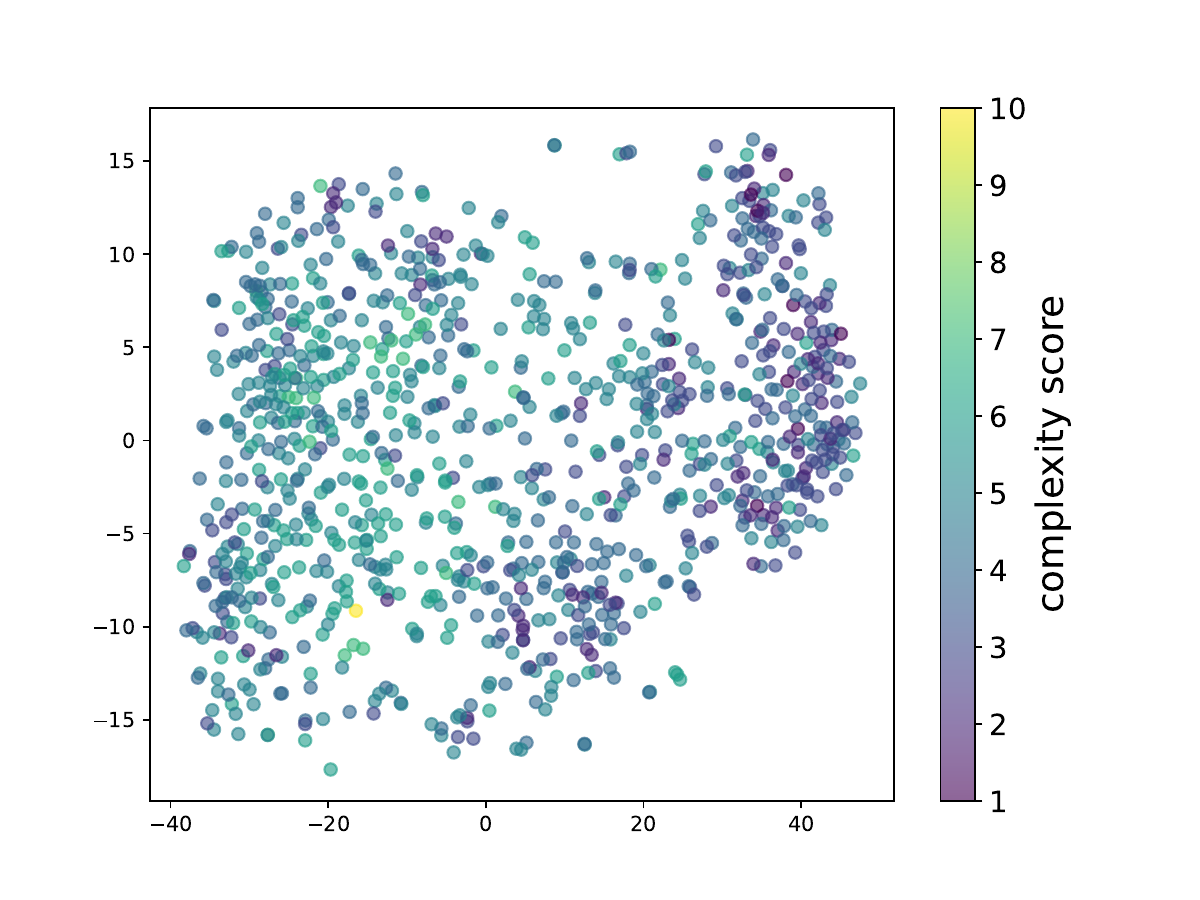}
		\subcaption{Complexity Coloring \\ CLIC}
		\label{figure1_2}
	\end{minipage}
	\begin{minipage}{0.49\linewidth}
		\centering
		\includegraphics[width=1\linewidth]{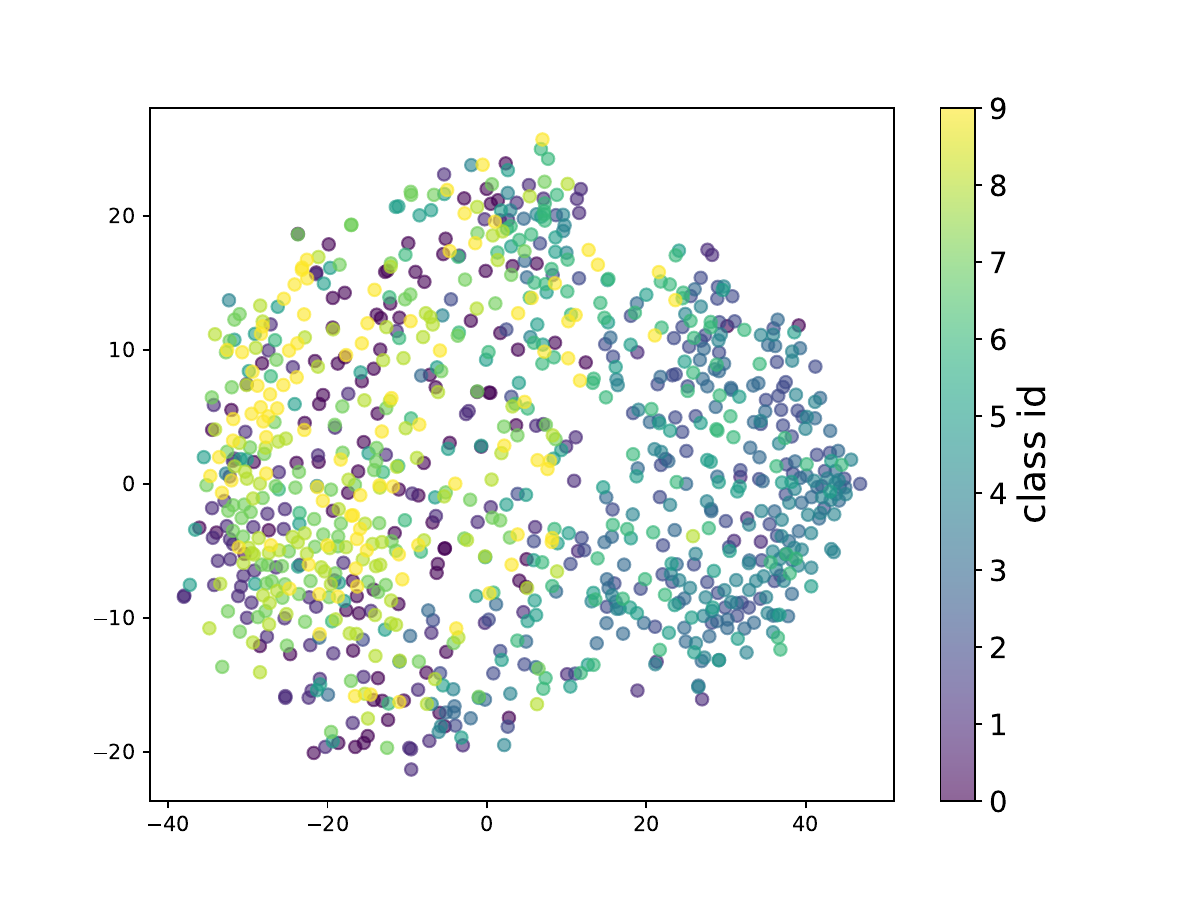}
		\subcaption{Class Coloring \\ CLICv2}
		\label{figure1_3}
	\end{minipage}
	\begin{minipage}{0.49\linewidth}
		\centering
		\includegraphics[width=1\linewidth]{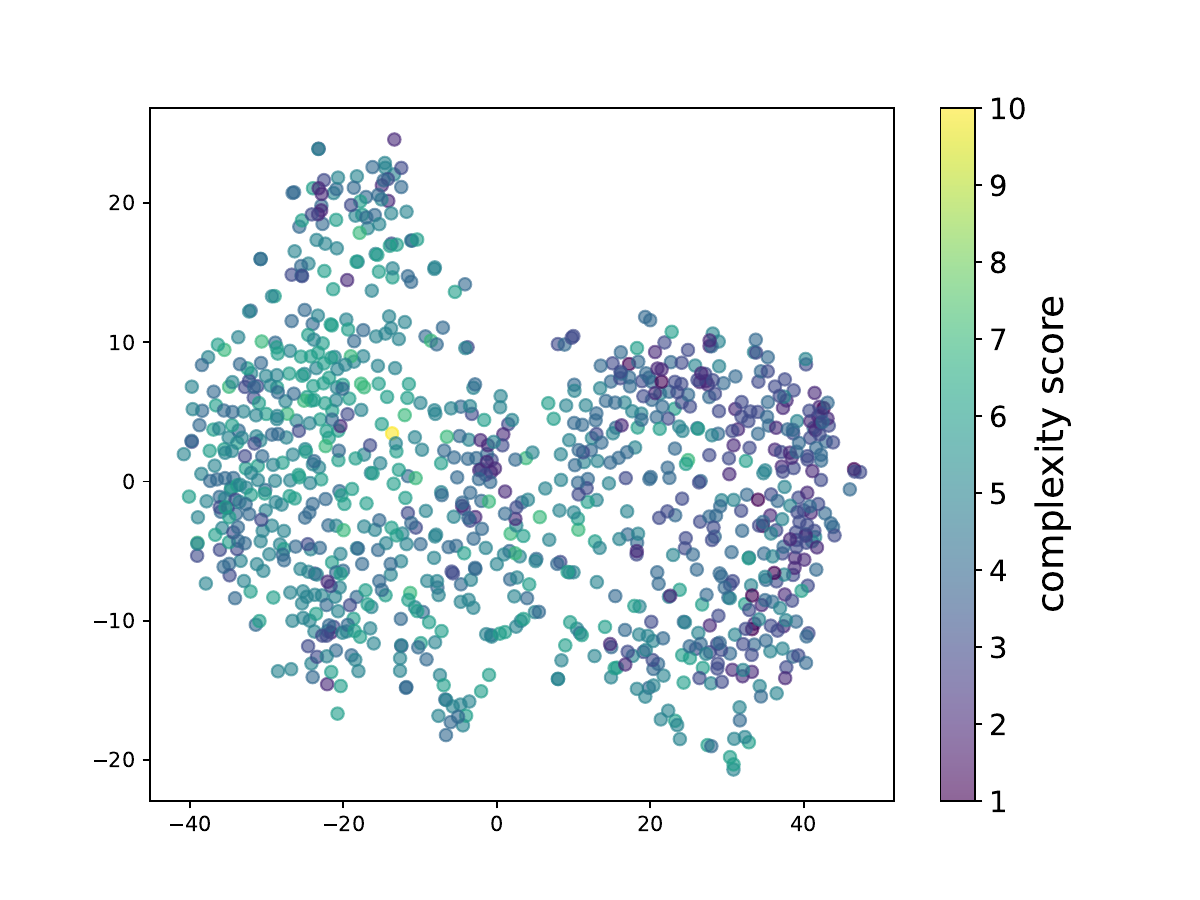}
		\subcaption{Complexity Coloring \\ CLICv2}
		\label{figure1_4}
	\end{minipage}
\caption{\textbf{\textit{t}-SNE visualization.} All feature are taken from the output of last layer in query encoder. The \textbf{complexity score} (b)(d) is divided into ten segments. The \textbf{class id} (a)(c) is from 0 to 9.}
\label{fig7}
\end{figure}

\textbf{\textit{t}-SNE.} \cref{fig7} gives the visualization results of CLIC and CLICv2 after \textit{t}-SNE~\cite{tsne} dimensionality reduction of image features. As shown in \cref{fig7}, CLIC (top row) still presents some clustering tendency in the class dimension, indicating that its feature distribution is partially affected by the image semantics; at the same time, it does not present a clear gradient differentiation when coloring by complexity. In CLICv2 (bottom row), on the other hand, samples of the same class are no longer clustered in the feature space, indicating that the model has successfully weakened the class interference to achieve content invariance; when colored by complexity, the samples form a more continuous gradient distribution, showing that CLICv2 is more discriminative for complexity learning. Thus, CLICv2 better suppresses image content information and focuses on complexity.

\section{Conclusion}

In this paper, we propose a CLICv2 framework based on content invariance constraints to address the problem of image semantic (class) information being easily introduced during the positive sample selection process in the traditional CLIC method, which affects the image complexity representation. We adopt the shifted patchify strategy to construct pairs of positive samples by shifting the random directions of patches and design the Patch-wise contrastive loss to achieve effective contrastive learning at the local level. In addition, the auxiliary task Masked Entropy Modeling is introduced to enhance further the model's ability to perceive the global complexity by reconstructing the information of the masked region with entropy as the goal. The experimental results fully demonstrate that CLICv2 performs well in the overall evaluation metrics and maintains high consistency across different classes of images, reflecting its strong content invariance.

{
    \small
    \bibliographystyle{ieeenat_fullname}
    \bibliography{main}
}

\end{document}